\pdfoutput=1
\documentclass[11pt]{article}

\usepackage{ACL2023}
\usepackage{times}
\usepackage{latexsym}
\usepackage{float}

\usepackage[T1]{fontenc}
\usepackage[utf8]{inputenc}

\usepackage{microtype}

\usepackage{inconsolata}

\usepackage{tikz} % latex figure
\usetikzlibrary{shapes,arrows} % latex figure
\usepackage{amsmath}
\usepackage{amssymb}
\usepackage{booktabs,multirow} 
\usepackage{tabularx}
\usepackage[ruled,vlined]{algorithm2e}
\usepackage{forloop}
\usepackage{array,multirow,graphicx}
\usepackage{makecell}

\def\forwardModel{F_{LLM}}
\def\backwardModel{B_{LLM}}
\def\methodName{InterrogateLLM}

 \title{InterrogateLLM: Zero-Resource Hallucination Detection in LLM-Generated Answers}

\author{
\begingroup
\begin{tabular}{c}
  Yakir Yehuda$^{*~1,2}$ \qquad 
  Itzik Malkiel$^{*~1}$   \qquad 
  Oren Barkan$^{3}$ \\   
  \qquad 
  Jonathan Weill$^{4}$ \qquad 
  Royi Ronen$^{1}$ \qquad 
  Noam Koenigstein$^{4}$
\end{tabular} 
\endgroup
\\ \\
$^1$Microsoft \qquad 
$^2$Technion \qquad 
$^3$ The Open University \qquad 
$^4$Tel-Aviv University
}

\begin{document}

\maketitle

\def\thefootnote{*}\footnotetext{Denotes equal contribution.}\def\thefootnote{\arabic{footnote}}

\begin{abstract}
Despite the many advances of Large Language Models (LLMs) and their unprecedented rapid evolution, their impact and integration into every facet of our daily lives is limited due to various reasons. One critical factor hindering their widespread adoption is the occurrence of hallucinations, where LLMs invent answers that sound realistic, yet drift away from factual truth. 
In this paper, we present a novel method for detecting hallucinations in large language models, which tackles a critical issue in the adoption of these models in various real-world scenarios.
Through extensive evaluations across multiple datasets and LLMs, including Llama-2, we study the hallucination levels of various recent LLMs and demonstrate the effectiveness of our method to automatically detect them. Notably, we observe up to 87\% hallucinations for Llama-2 in a specific experiment, where our method achieves a Balanced Accuracy of 81\%, all without relying on external knowledge
\footnote{Our code, datasets, and task prompts can be found \href{https://github.com/yakir-yehuda/InterrogateLLM}{here}. }.
\end{abstract}

\section{Introduction}
Human studies have shown that people tend to be inconsistent when they are not telling the truth~\cite{brewer1999beliefs}. As such, a common interrogation technique consists of repeated interviews that attempt to challenge the interviewer's consistency in order to assess its credibility \cite{granhag2001deception}. Truth tellers' answers are well-grounded in their memory, hence, inconsistencies in the respondent's answers are a strong indication of her not telling the truth~\cite{brewer1999beliefs,dianiska2023effect}. Inspired by these studies, we present a novel method for hallucination detection in LLMs. Our approach, which we call 
\methodName,
employs a systematic evaluation of model-generated responses for potential hallucinations by repeating the process of reconstructing a query from its generated answer.

Repeated interviews are a very common and effective verification technique for human interrogations, however, it is not foolproof. In some cases, respondents manage to provide repeated false states that are consistent, while in other cases, truth-tellers may provide inconsistent responses due to memory errors~\cite{bartlett1995remembering}. 
In a similar fashion, our method is not flawless; it represents an additional step towards addressing the yet unsolved problem of hallucination detection.
Nevertheless, similar to the use of consistency tests in humans, commonly employed for their effectiveness, our method also demonstrates high efficacy.

In recent years, the emergence of LLMs such as GPT-3 \cite{Tom_Brown_few_shot}, PaLM \cite{chowdhery2022palm}, and LLama \cite{llama1,llama2} has revolutionized natural language processing. These models enable machines to understand and generate human-like text with unprecedented fluency and coherence.
Trained on vast amounts of text data, they have demonstrated remarkable capabilities in various applications, from automated content generation to virtual assistants, and beyond. However, their remarkable performance comes with a set of challenges and concerns that need to be addressed for their responsible and effective use.
A major concern is the phenomenon of hallucination, whereby these language models generate misleading, potentially harmful, or erroneous text. Hallucination can be characterized by the presence of false information in the output generated by the language model that lacks a factual basis. There are significant challenges associated with the deployment of large language models in real-world applications, especially in those involving critical information or decision-making processes.

Detecting and minimizing hallucinations in LLMs is crucial for ensuring their trustworthiness and reliability, especially in contexts where these models play a pivotal role in communication and decision-making. Existing methods for evaluating model-generated text often rely on surface-level metrics such as fluency and coherence, which may not effectively capture the underlying issue of hallucinations. Therefore, there is a pressing need for a systematic and effective method to detect and mitigate hallucinations in the outputs of these models. Despite its significance, addressing this challenge remains an open issue \cite{ji2023}.

Our method, \methodName{}, operates on the premise that language models exhibiting hallucinations produce inconsistent and incorrect responses to subsequent queries based on the hallucinated information. 
To identify hallucination in a generated answer, our approach involves prompting the model multiple times to reconstruct the input query using the generated answer. Subsequently, \methodName{} quantifies the inconsistency level between the original query and the reconstructed queries.
By leveraging the observed inconsistencies, our approach effectively identifies potential instances of hallucination.
When a large-language model generates a hallucination, it struggles to consistently reconstruct the original query, leading to variations in responses. This interrogation strategy serves as the cornerstone of our approach for detecting hallucinations in generated answers.

The contributions of our paper are outlined as follows: (1) introduction of the \methodName{} method designed for detecting hallucinations in textual answers generated by LLMs. (2) we propose an innovative evaluation approach specifically tailored to the task of hallucination detection, leveraging three datasets associated with our proposed text generation tasks. (3) we investigate the hallucination levels exhibited by recent LLMs, including Llama2, shedding light on their fidelity levels. (4) we present comprehensive performance reports on \methodName{} and its variants, conducting a thorough comparison with alternative methods through extensive evaluations.

\begin{figure*}[t]
\includegraphics[width=1.0\linewidth]{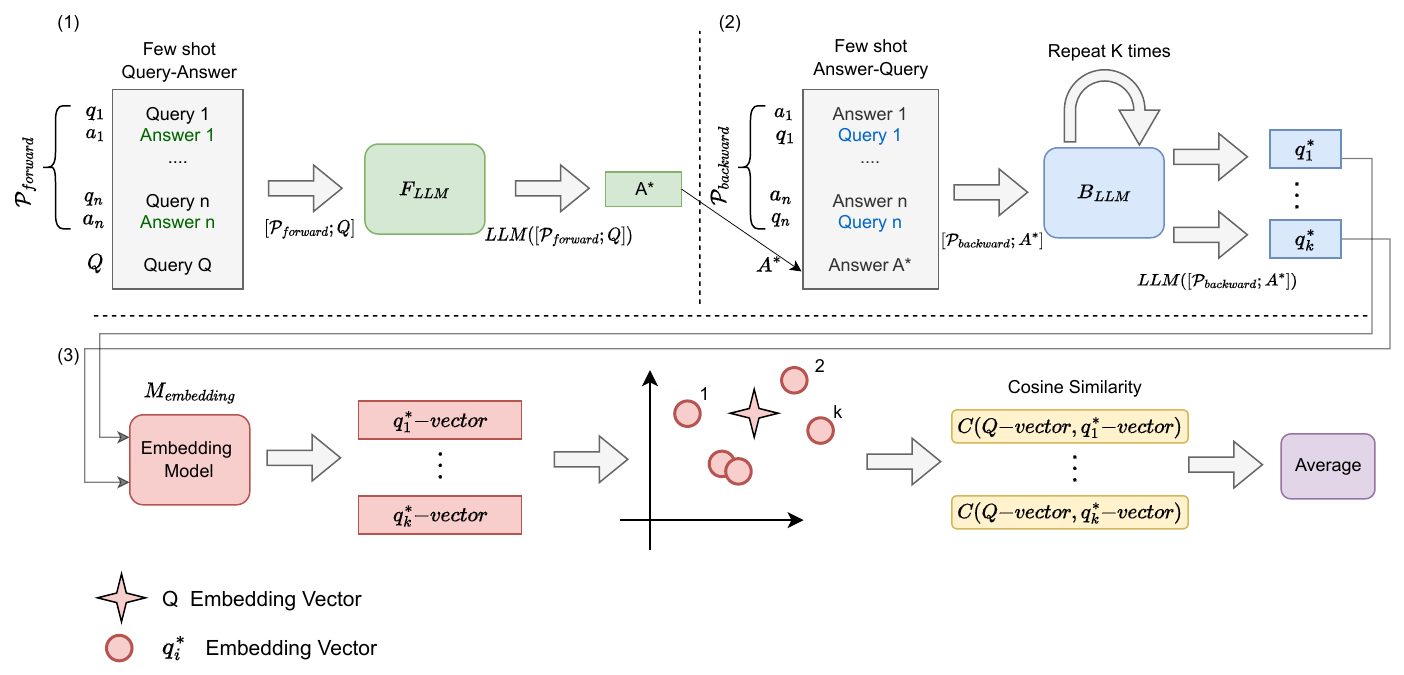}
\centering
\caption{An illustration of the \methodName{} method. (1) A few-shot prompt and a query are fed into $\forwardModel$, which generates an answer. (2) The shots in the prompt are then reversed, forming a sequence of answer-question pairs, with the generated answer constructed on top. The $\backwardModel$ is then used to generate $K$ queries that correspond to the generated answer. Ideally, the generated queries should recover the original query from the forward phase. (3) The set of recovered questions is then embedded by a language model and compared with the original question, producing a final score that determines whether the generated answer suffers from hallucination.}
\label{fig:method}
\end{figure*}

\section{Related Work}
Hallucinations have been %extensively 
explored in various natural language generation tasks, including translation, summarization \cite{kryscinski-etal-2020-evaluating,maynez-etal-2020-faithfulness}, dialogue generation \cite{shuster-etal-2021-dialogue}, and question-answering \cite{lin-etal-2022-truthfulqa}. This is well-documented in a recent comprehensive survey conducted by \cite{ji2023}, which provides an insightful overview of hallucinations in diverse natural language generation contexts.

In \cite{liu-etal-2022-token}, the authors presented a token-level reference-free hallucination detection task along with an additional dataset designed for hallucination detection in free-form text. This dataset consists of textual passages with perturbations, and the objective is to determine whether the entire passage exhibits hallucinations. It is crucial to emphasize that our task differs from their setup, as we specifically address hallucination detection within few-shot prompts involving query-answer sequences.

To address inconsistencies in generated text, SelfCheckGPT, introduced by \citet{manakul2023selfcheckgpt}, leverages multiple stochastic samples generated by LLMs using the same query. SelfCheckGPT evaluates the coherence between the response and the stochastic samples by querying the same LLM multiple times. Specifically, it incorporates an additional prompt that includes a stochastic sample and a sentence from the generated text and predicts whether the sentence is supported by the stochastic sample. 
The approach validates each sentence by conditioning the LLM on each stochastic sample.
The methodology of SelfCheckGPT encompasses various approaches, including one based on BERTScore \cite{fu2023gptscore}, and another employing a multiple-choice question answering and generation approach (MQAG) \cite{manakul2023mqag}, as well as n-gram and LLM-Prompting. 
Our method is benchmarked against this baseline, using the last approach in our study.

In recent research, \citet{azaria2023internal} proposed a method employing a multilayer perceptron classifier that uses hidden representations from language models to predict sentence truthfulness. However, this approach necessitates labeled data for supervised training and access to the internal states of the language model, which may not always be readily available.
In \cite{kadavath2022language}, the authors present a self-evaluation technique where models are trained to predict their knowledge of the answer to any given free-form question. This approach entails prompting the language model to internally assess the accuracy of its previous predictions, including estimating the likelihood that its generated response or answer is correct. It is worth noting that this method requires labeled data for model training, making it a supervised task, which differs from our settings.

\section{Problem setup}
We assume a source domain of textual queries and a target domain of textual answers. A few-shot prompt\footnote{While our approach assumes a provided few-shot prompt, it stays adaptable to many zero-shot tasks where the creation of few-shot prompts is feasible.} \cite{Tom_Brown_few_shot}, 
a corresponding query $Q$ and a LLM denoted by $\forwardModel$, are provided. The query is constructed on top of the prompt and fed into the LLM to generate an answer to the query. Our task is to detect whether the generated answer suffers from hallucinations.

The few-shot prompt is constructed as a sequence of query-answer pairs. The pairs are denoted by $\left\{\left(q_{i}, a_{i}\right)\right\}_{i=1}^n$, where $q_{i}$ represents a query and $a_{i}$ its corresponding answer. The prompt can be expressed as follows:
\begin{equation}
   \mathcal{P}_{forward} = q_{1}, a_{1}, \ldots, q_{n}, a_{n}
\end{equation}

The $\forwardModel$ is queried with the concatenation of the query $Q$ on top of the prompt $\mathcal{P}_{forward}$, which retrieves a generated answer denoted by $A^*$, signifying the response to the query $Q$.
In other words, the prompt $\mathcal{P}_{forward}$ and the query $Q$ are fed into the LLM as follows:
\begin{equation}
   A^* = \forwardModel([\mathcal{P}_{forward}; Q])
\end{equation}
Our task is to determine whether the generated answer $A^*$ exhibits hallucinatory content.

\section{The \methodName{} method}

In our approach, we introduce a backward process for reconstructing the original query $Q$ from the generated answer $A^*$. We create a new prompt by reversing the given prompt $\mathcal{P}_{forward}$.
The reversed prompt rearranges the order of the query-answer pairs to pairs of answer-query. The reversed prompt, denoted as $\mathcal{P}_{backward}$, can be expressed as follows:
\begin{equation}
   \mathcal{P}_{backward} = a_{1}, q_{1}, \ldots, a_{n}, q_{n}
\end{equation}

The generated answer $A^*$ is then concatenated to the end of the reversed prompt $\mathcal{P}_{backward}$, and the entire sequence is passed either by the same LLM defined above, by a different LLM, or by an ensemble of LLMs. For ease of reference and clarity, we collectively refer to the LLMs involved in this step as $\backwardModel$. In other words, in this process, we map the generated answer to the source domain, by querying one or more LLMs, each trying to reconstruct the original query $Q$. 
By denoting the set of reconstructed queries as $Q^*$, this \textit{``backward''} process can be expressed as:
\begin{equation}
   Q^* = \backwardModel([\mathcal{P}_{backward}; A^*])
\end{equation}
Note that the size of $Q^*$ depends on the number of LLMs used in $\backwardModel$.

The motivation for employing a backward process is to reconstruct the original query $Q$ based on the generated answer $A^*$. If the initial LLM suffers from hallucinations during the generation of $A^*$, then $A^*$ may drift from the correct answer to $Q$. Consequently, a backward process operating on $A^*$ is prone to deviating from $Q$ on the way back. In other words, in the case of hallucination in $A^*$, the set of reconstructed queries $Q^*$ is likely to diverge from the original query $Q$.

In \methodName{}, this backward process is repeated multiple times ($K$ times for each model in $\backwardModel$, see Sec.~\ref{sec:impel} for more details), with variable temperature values, as explained below. Therefore,
% $|Q^*| = K * |\backwardModel|$.
\begin{align*}
    |Q^*| = K * |\backwardModel|
\end{align*}

\begin{algorithm}[t]
\caption{Hallucination Detection}
\label{algo:qa_verification}
\small
% \resizebox{0.6\linewidth}{!}{%
\LinesNumbered
\SetKwComment{Comment}{/* }{ */}
 % \Comment*[r]{This is a comment}
\KwIn{$Q$, $\forwardModel$, $\backwardModel$, $M_{embedding}$, $\left\{\left(q^{(i)}, a^{(i)}\right)\right\}_{i=1}^n$, $\tau$}
\KwOut{True if the generated answer is an hallucination, False otherwise}

\textbf{Step 1: Query Forward Pass}\
$\mathcal{P}_{forward} = q_{1}, a_{1}, \ldots, q_{n}, a_{n}$ \
$A^* = \forwardModel([\mathcal{P}_{forward}; Q])$\

\textbf{Step 2: Query Reconstruction}\
$\mathcal{P}_{backward} = a_{1}, q_{1}, \ldots, a_{n}, q_{n}$ \

$Q^* = \{\}$\

\For{$i=1$ \KwTo $K$}{
    \textbf{Substep 1: Reverse Pass}\
    $Q^* = Q^* \cup \backwardModel([\mathcal{P}_{backward}; A^*])$\    
    % \textbf{Substep 3: Cosine Similarity}\
    % $Q^*_{sim} = \{C(q^*_{vec}, Q_{vec}): \forall q^*_{vec} \in Q^*_{vec} \}$\ 
}
\textbf{Step 3: Text Embedding}\
$Q_{vec}=M_{embedding}(Q)$\
$Q^*_{vec} = \{M_{embedding}(q^*): \forall q^* \in Q^* \}$\ 
    
\textbf{Step 4: Verification}\
$\text{sim}(Q, Q^*) = \underset{\forall q^*_{\text{vec}} \in Q^*_{\text{vec}}}{\text{AVG}}\left(\left[C(Q_{\text{vec}}, q^*_{\text{vec}})\right]\right)$

\eIf{$\text{sim}(Q, Q^*) \geq \tau$}{
  \Return{False}\
}{
  \Return{True}\
}
% }
\end{algorithm}

To determine if $A^*$ suffers from hallucination, a language embedding model is utilized to assess the similarity between the set of reconstructed queries $Q^*$ and the original query $Q$. Both the generated queries and the original query are transformed into vectors within the same embedding space. For a given embedding model $M_{embedding}:text \rightarrow \mathbb{R}^{D}$, which generates $D$-dimensional vectors from the input text, the embedding vector for the original query $Q$ is denoted as $Q_{vec} = M_{embedding}(Q)$. Similarly, the embedding vectors for the generated queries are denoted by:
% $Q^*_{vec} = M_{embedding}(Q^*)$.
% $Q^*_{vec} = \{M_{embedding}(q^*): \forall q^* \in Q^* \}$.
\begin{align*}
    Q^*_{vec} = \{M_{embedding}(q^*): \forall q^* \in Q^* \} 
\end{align*}
Subsequently, the cosine similarity between the embedding vectors of the predicted queries $Q^*_{vec}$ and the original query $Q_{vec}$ is calculated as follows:
\begin{equation}
\label{eq:avg}
\text{sim}(Q, Q^*) = \underset{\forall q^*_{\text{vec}} \in Q^*_{\text{vec}}}{\text{AVG}}\left(\left[C(Q_{\text{vec}}, q^*_{\text{vec}})\right]\right)
\end{equation}
Here, $C$ represents the cosine similarity function: 
\begin{equation}
C(u, v) = \left(\frac{u \cdot v}{\|u\|\|v\|}\right)
\end{equation}
for $u,v \in \mathbb{R}^{D}$, where $D$ is the dimension of the vectors.
In other words, the cosine similarity is calculated for each $q^*_{\text{vec}}$ in the set $Q^*_{\text{vec}}$, and the results are then averaged to obtain the final similarity score.

Finally, \methodName{} predicts hallucinations if the similarity score exceeds a predetermined threshold $\tau$. In essence, when the reconstructed queries exhibit a significant divergence from the original query, \methodName{} signifies that there is a potential hallucination in $A^*$. More details about the selection of $\tau$ can be found in Sec.~\ref{sec:impel}. The \methodName{} method is illustrated in Fig.~\ref{fig:method}, and outlined in Alg.~\ref{algo:qa_verification}.

\subsection{Variable temperatures}
We introduce an exploratory extension into \methodName{}, exploring the impact of diverse temperature values on the accuracy of the detections.
In standard LLMs, the temperature parameter influences the likelihood of selecting the next token during the answer generation process. A higher temperature (e.g., 1.0) makes the output more \emph{creative} and \emph{diverse}, while a lower temperature (e.g., 0.2) makes the output more focused and deterministic. Specifically, the temperature is applied through a softmax function that transforms a vector into a probability distribution. 
In text generation, the softmax function is applied to the model's logit vector, which corresponds to the supported tokens in the vocabulary.
The softmax operation can be written as follows:
\begin{equation}
    P_i = \frac{e^{{z_i}/{T}}}{\sum_{j=1}^{N} e^{{z_j}/{T}}}
\end{equation}
Where $P_i$ is the probability of selecting the $i$-th token in the vocabulary, $z$ is the logit vector, $T$ is the temperature parameter and $N$ is the number of tokens in the vocabulary. When $T$ is high (low), the exponential function ${e^{{z_i}/{T}}}$ is less (more) sensitive to small differences in the logit values, making the probabilities more diverse (focused). 

As complementary experimental explorations, we examine the influence of temperature values on \methodName{} during the backward process, which is iterated $K$ times. By introducing dynamic temperature adjustments, our goal is to study the method's accuracy when employed with a range of backward processes exhibiting diverse creativity levels. To this end, we set the temperature for each backward process as follows:
\begin{equation}
\label{eq:temperature}
T_i = T_0 + \frac{1.0 - T_0}{K} \cdot i
\end{equation}
where $T_i$ represents the temperature for the $i$-th backward pass ($0 \leq i < K$), and $T_0$ is the model default temperature (see Sec.~\ref{sec:impel} for more details). 

This temperature scheduling allows for facilitating a controlled ascent in temperatures across the multiple backward processes, promoting enhanced exploration in the space of reconstructed queries.
The details and results of this additional study are reported in the experiments, Sec.~\ref{sec:ablation}.

\begin{table}[ht!]
\centering
\begin{tabularx}{0.99\linewidth}{l *{3}{>{\centering\arraybackslash}X}}
\toprule
& \multicolumn{3}{c}{\textbf{Hallucination Rate}} \\
\cmidrule(lr){2-4}
\textbf{$\forwardModel$} & Movies & Books & GCI \\
\midrule
GPT3  & 37\% & 38\% & 0\%   \\
Llama-2 (7B) & 87\% & 66\% & 25\% \\
Llama-2 (13B) & 72\% & 58\% & 60\% \\
\bottomrule
\end{tabularx}
\caption{Hallucination rates for each dataset and $\forwardModel$.}
\label{Tab:hallucination_rate}
\end{table}

\begin{table*}[t!]

\centering
\resizebox{0.78\textwidth}{!}{
\begin{tabular}{@{}l@{~}
p{0.175\textwidth}<{\centering\arraybackslash}l@{~}p{0.2\textwidth}
p{0.05\textwidth}<{\centering\arraybackslash}c@{~}|
p{0.05\textwidth}<{\centering\arraybackslash}c@{~}|
p{0.05\textwidth}<{\centering\arraybackslash}c@{~}}
% {@{}l@{~}l@{~}l@{~}l@{~}c@{~}c@{~}|c@{~}c@{~}|c@{~}c@{~}}
\toprule 
& & & & \multicolumn{2}{c}{Movies}& \multicolumn{2}{c}{Books}& \multicolumn{2}{c}{GCI}\\
\cmidrule(lr){5-6}
\cmidrule(lr){7-8}
\cmidrule(lr){9-10}

\textbf{$\forwardModel$ }& \multicolumn{3}{c}{Method} & 
\textbf{AUC}& \textbf{B-ACC} &
\textbf{AUC}& \textbf{B-ACC} &
\textbf{AUC}& \textbf{B-ACC} \\
\midrule

%% GPT 3 results
\multirow{7}{*}{\rotatebox{90}{\textbf{GPT3}}} 
& \multirow{4}{*}{$\methodName$} 
% & \multirow{4}{*}{\rotatebox{90}{\textbf{$\methodName$}}}
& \multirow{4}{*}{\rotatebox{90}{\textbf{$\backwardModel$}}}

& \multirow{1}{*}{GPT3} 
&0.817 &\textbf{0.739}  % movies
&0.709 &\textbf{0.673}  % books
&- & \textbf{0.994} \\

& & & \multirow{1}{*}{Llama-2 (7B)}
&0.751 &0.639  % movies
&0.646 &0.616  % books
&- & 0.983 \\

& & & \multirow{1}{*}{Llama-2 (13B)}
&0.789 &0.695  % movies
&0.684 &0.640  % books
&- & 0.983 \\  % gci

& & & \multirow{1}{*}{Ensemble}
&\textbf{0.818} &0.699  % movies
&\textbf{0.710} &0.656  % books
&- & 0.983  \\   % gci

\cmidrule{2-10}

& \multicolumn{3}{c}{SBERT-cosine} 
&0.616 &0.500  % movies
&0.534 &0.500  % books
&- & 0.550  \\   % gci

& \multicolumn{3}{c}{ADA-cosine} 
&0.709 &0.500  % movies
&0.530 &0.500  % books
&- & 0.591  \\     % gci

& \multicolumn{3}{c}{SelfCheckGPT} 
&0.782 &0.684  % movies
&0.685 &0.629  % books
&- & 0.977  \\     % gci

\midrule
\midrule
\multirow{7}{*}{\rotatebox{90}{\textbf{Llama-2 (7B)}}} 
& \multirow{4}{*}{$\methodName$} 
% & \multirow{4}{*}{\rotatebox{90}{\textbf{$\methodName$}}}
& \multirow{4}{*}{\rotatebox{90}{\textbf{$\backwardModel$}}}

& \multirow{1}{*}{GPT3} 
&0.824 &0.786 % movies
&\textbf{0.828} &\textbf{0.787}  % books
&0.965 & 0.952 \\

& & & \multirow{1}{*}{Llama-2 (7B)}
&0.823 &0.750 % movies
&0.761 &0.707 % books
&0.959 & 0.958 \\

& & & \multirow{1}{*}{Llama-2 (13B)}
&0.828 &0.775  % movies
&0.795 &0.734  % books
&\textbf{0.969} & \textbf{0.960}  \\ % gci

& & & \multirow{1}{*}{Ensemble}
&\textbf{0.874} &\textbf{0.813}  % movies
&0.822 &0.761  % books
&0.951 & 0.948   \\   % gci

\cmidrule{2-10}

& \multicolumn{3}{c}{SBERT-cosine} 
&0.586 &0.516  % movies
&0.552 &0.486  % books
&0.957 &0.548  \\     % gci

& \multicolumn{3}{c}{ADA-cosine} 
&0.770 &0.501  % movies
&0.641 &0.499  % books
&0.950 & 0.820  \\     % gci

& \multicolumn{3}{c}{SelfCheckGPT} 
&0.820 &0.634  % movies
&0.784 &0.710  % books
&0.963 & 0.927  \\     % gci

\midrule
\midrule
\multirow{7}{*}{\rotatebox{90}{\textbf{Llama-2 (13B)}}} 
& \multirow{4}{*}{$\methodName$} 
% & \multirow{4}{*}{\rotatebox{90}{\textbf{$\methodName$}}}
& \multirow{4}{*}{\rotatebox{90}{\textbf{$\backwardModel$}}}

& \multirow{1}{*}{GPT3} 
&0.806 &0.753  % movies
&0.804 &\textbf{0.754}  % books
&0.989 & 0.982 \\

& & & \multirow{1}{*}{Llama-2 (7B)}
&0.788 &0.706  % movies
&0.742 &0.697  % books
&\textbf{1.000} & \textbf{1.000} \\

& & & \multirow{1}{*}{Llama-2 (13B)}
&0.801 &0.746  % movies
&0.771 &0.709  % books
&0.995 & 0.991  \\ % gci

& & & \multirow{1}{*}{Ensemble}
&\textbf{0.842} &\textbf{0.773}  % movies
&\textbf{0.807} &0.733  % books
&0.992 & 0.964  \\    % gci

\cmidrule{2-10}

& \multicolumn{3}{c}{SBERT-cosine} 
&0.539 &0.505  % movies
&0.573 &0.497  % books
&0.955 &0.546 \\

& \multicolumn{3}{c}{ADA-cosine} 
&0.728 &0.500  % movies
&0.600 &0.500 % books
&0.966 &0.852  \\ % gci

& \multicolumn{3}{c}{SelfCheckGPT} 
&0.794 & 0.689  % movies
&0.751 & 0.693  % books
&0.934 & 0.891  \\     % gci                

\bottomrule
\end{tabular}
}
\caption{Hallucination detection results for all models and datasets. \methodName{} is reported with $K=5$ and variable temperature values. For each dataset and $\forwardModel$, we compare \methodName{} and its variants to all other baselines. As GPT3 does not suffer from hallucinations on the GCI dataset, only the ACC metric is reported (in the B-ACC column).
}
\label{Tab:main}
\end{table*}

\section{Experiments}

To assess the efficacy of \methodName{} in detecting hallucinations, and due to the absence of prior datasets for hallucination detection in few-shot prompt settings, we adapted three public datasets. For each dataset, we designed a text generation task along with a verification process to ensure the accuracy of the generated answers.
The verification is implemented by employing simple heuristic functions that exploit additional information that is present in the datasets.
During the evaluation of hallucination detection methods, the detection predictions are compared against the verification results. 
Importantly, the \methodName{} method operates independently of any external knowledge, making it versatile and applicable to a broad spectrum of tasks.

\subsection{Datasets and Tasks}
A comprehensive experimental evaluation was conducted using three different datasets to thoroughly evaluate our hallucination detection method across various domains. All three datasets provide a multifaceted evaluation of our technique, revealing its versatility across various types of information and content and allowing us to test the robustness of our hallucination detection method across a wide range of datasets and domains. 

\subsubsection{The Movies Dataset}
The Movies 
Dataset\footnote{\href{https://www.kaggle.com/datasets/rounakbanik/the-movies-dataset}{The Movies Dataset}} is a collection of movie-related data that is publicly available for analysis and research. The dataset contains a variety of details about movies that were released before July 2017. 
The dataset includes 26 million ratings and 750,000 tag applications for all 45,000 movies provided by 270,000 users. 

A subset of 3000 samples with movie titles and release years associated with the movie cast was sampled from the Movies dataset. The task is to predict the cast of a movie based on the movie's name and release year. The few-shot prompt contains a few examples mapping a movie's name and release year to its cast. The prompt is in the following format: "Query: What actors played in the $x$ movie $y$?" where $x$ is the release year and $y$ is the movie name. 
Cast members' full names are expected in answers, and ground truth labels use Intersection Over Union (IOU) scores, considering any IOU score below 80\% as a hallucination.

\subsubsection{Books Dataset}
The second dataset ("books dataset")\footnote{\href {https://www.kaggle.com/datasets/saurabhbagchi/books-dataset}{Books Dataset}} is derived from Amazon and includes over 200,000 literary books. This public dataset provides an overview of diverse literary books available on the Amazon platform. Each record includes details like book title, authors, publishers, and publication year. 

We sampled a subset of 3,000 samples, including titles, dates, authors, and publishers. The task is to predict the author and publication year based on the book title. The prompts are structured as "Who is the author of the book $x$, what year was it published?", where $x$ is the book title. The ground truth is established by checking for a match between the elements (author name, release year) in the answer. 

\subsubsection{Global Country Information (GCI)}
The ``Global Country Information''\footnote{\href {https://www.kaggle.com/datasets/nelgiriyewithana/countries-of-the-world-2023}{GCI Dataset}} (GCI) is a public dataset containing information on 181 countries. Detailed information about each country is provided, including its name, land area, capital or major city, GDP, and more. This dataset offers a comprehensive representation of global country information. 
In the GCI dataset, we concentrate on country and capital pairs. The task involves determining a country's capital by asking, "What is the capital of $x$?"

Samples from the above three datasets can be found in the supplementary Sec.~\ref{sec:samples}. The prompts used in each dataset and the reversed prompts created by \methodName{}, can be found in the code\footnote{ \href{https://github.com/yakir-yehuda/InterrogateLLM}{GitHub project}}.

\subsection{Implementation details}
\label{sec:impel}

We set $K=5$ and $\tau = 0.91$ %consistently
across all experiments. Maintaining a relatively small value for $K$ facilitates rapid benchmarking of various models on datasets in our evaluations, that encompass tens of thousands of generated answers.
The hyperparameter $\tau$ was determined through an analysis of ada002 embeddings on a third-party dataset. This involved embedding both similar and dissimilar sentence pairs within the QQP dataset~\cite{chen2018quora} and selecting the optimal threshold that effectively distinguished between the two distributions. 
The initial temperature $T_0$ was set to the default temperature of each of the evaluated LLMs, specifically $0.6$ for GPT3 and both Llama-2 models.
The embedding model used in \methodName{} leverages the latest OpenAI's model, ada002\footnote{https://platform.openai.com/docs/guides/embeddings/use-cases}.

In our experiments, we used one A100 GPU. A single application of \methodName{} with the full method for $k=1$, using an ensemble of three models, takes up to 2 seconds. Consequently, benchmarking \methodName{} across the three datasets takes up to $\sim3.44$ hours. 
Further insights into the hyperparameters and experimental environment will be detailed in the following subsections.

\subsection{Baselines}
We compare our method with the following baselines, evaluated on all datasets and $\forwardModel$ models:

\noindent{\textbf{SBERT-cosine:}}in this baseline, we employ a pre-trained SBERT model~\cite{reimers-gurevych-2019-sentence} to embed both the query and the generated answer. We then calculate the cosine similarity between them and predict "hallucination" if the similarity falls below a threshold $SBERT_\tau$. The threshold was determined by using the same process described in Sec.\ref{sec:impel}, this time with SBERT embeddings.

\noindent{\textbf{ADA-cosine:}} similar to SBERT-cosine but employs the recent openAI model ada002. The value of $\tau$ used here is consistent with the one in Sec.\ref{sec:impel}.

\noindent{\textbf{SelfCheckGPT with Prompt:}} utilizes the same $\forwardModel$ in each task, SelfCheckGPT generates additional $N$ stochastic LLM response samples, denoted as ${S_1, S_2, ..., S_n}$, using the same query. 
Then, it scores the consistency between the generated response and the stochastic samples, by querying an LLM to determine whether the $i$-th sentence in $A^*$ is supported by the corresponding sample $S_i$.
The final inconsistency score is computed by averaging the sentence scores. In the experiments, this scoring step is evaluated using GPT-3 for all tasks.

\subsection{The hallucination rates}
We evaluate \methodName{} on answers generated by three recent LLMs for each of the datasets and tasks described above. The LLMs we evaluate are: GPT-3 \cite{Tom_Brown_few_shot} and Llama-2 \cite{llama2} (7b and 13b models).
Interestingly, in Tab.~\ref{Tab:hallucination_rate} we report the hallucination rates in the generated answers of the three models across the different datasets. 
Notably, GPT-3 exhibits a lower hallucination rate across all datasets and tasks, compared to the Llama-2 models. 

% Method Evaluation: 
\subsection{Hallucination detection results}
Binary predictions (hallucinations or not) are compared to the ground truth test labels of each dataset.

For each dataset and task, we employ \methodName{} with four different LLM choices for the backward step: GPT-3, Llama-2 (7B), and Llama-2 (13B), either individually or as ensembles of all three models.
In Tab. \ref{Tab:main}, we report the area under the curve (AUC) of the receiver operating characteristic and balanced accuracy (B-ACC) metrics.

As can be seen in the table, the different variants of our method improve upon all the baselines by a sizeable margin. 
Importantly, we note sizeable improvements also in comparison to SelfCheckGPT. This advantage attributed to \methodName{} stems from predicting the query back using the few-shot samples provided in the prompt, a factor entirely overlooked by SelfCheckGPT. Additionally, we observed that in many instances of hallucinations, the stochastic samples generated by SelfCheckGPT also exhibited the same mistake. Therefore, the SelfCheckGPT algorithm erroneously predicted the hallucinated $A^*$ as factual truth. 
This emphasizes the importance of our unique backward validation strategy, which differs from the query that initially caused the hallucination.
Within the variants of the \methodName{} method, we observe that the use of an ensemble in the backward process exhibits sizeable strides across the board, suggesting that model diversity can compensate for individual model weaknesses (see also Sec.\ref{sec:limitations}).

% movies k=1 - k=5 results
\begin{table*}[ht!]

\centering
\resizebox{0.85\textwidth}{!}{

\begin{tabular}{@{}l@{~}l@{~}c@{~}c@{~}c@{~}c@{~}c@{~}c@{~}c@{~}c@{~}c@{~}c@{~}c@{~}c@{}}
\toprule 
& & \multicolumn{2}{c}{k=1}& \multicolumn{2}{c}{k=2}& \multicolumn{2}{c}{k=3} & \multicolumn{2}{c}{k=4}& \multicolumn{2}{c}{k=5}\\
\cmidrule(lr){3-4}
\cmidrule(lr){5-6}
\cmidrule(lr){7-8}
\cmidrule(lr){9-10}
\cmidrule(lr){11-12}
\textbf{$\forwardModel$ }&  \textbf{$\backwardModel$ }& 
\textbf{\  AUC\ \ }&  \textbf{\ B-ACC\ \ }&
\textbf{\  AUC\ \ }&  \textbf{\ B-ACC\ \ }&
\textbf{\  AUC\ \ }&  \textbf{\ B-ACC\ \ }&
\textbf{\  AUC\ \ }&  \textbf{\ B-ACC\ \ }&
\textbf{\  AUC\ \ }&  \textbf{\ B-ACC\ \ }\\
\midrule

%% GPT 3 results
\multirow{4}{*}{\textbf{GPT3}}
% \multirow{4}{*}{\rotatebox{90}{\textbf{GPT3}}}

& GPT3 &0.755 &0.710 
            &0.773 &\textbf{0.722} 
            &0.782 & 0.719 
            &0.786 &0.720 
            &\textbf{0.790} & 0.721  \\

& Llama-2 (7B) &0.701 &0.633 
                    &0.721 &\textbf{0.641} 
                    &0.727 &0.635 
                    &0.732 &0.638 
                    &\textbf{0.734} &0.631 \\

& Llama-2 (13B) &0.756 &0.688 
                    &0.772 &0.696 
                    &0.779 &\textbf{0.698} 
                    &0.783 &0.696
                    &\textbf{0.787} &0.697 \\

 & Ensemble & 0.796 & 0.690 
                &0.803 &0.694 
                &0.811 &0.694 
                &0.814 &0.695 
                &\textbf{0.815} &\textbf{0.695} \\
\midrule

%% Llama-2 (7B) results
\multirow{4}{*}{\textbf{Llama-2 (7B)}}
% \multirow{4}{*}{\rotatebox{90}{\textbf{Llama-2 (7B)}}}

& GPT3  &0.775 &0.774 
            &0.786 &0.778 
            &0.788 & 0.776
            &0.794 & \textbf{0.782} 
            &\textbf{0.798} & 0.780  \\

& Llama-2 (7B) &0.798 &0.754 
                    &0.815 &0.766 
                    &0.825 &0.757 
                    &0.831 &0.760 
                    &\textbf{0.830} &\textbf{0.766}  \\

& Llama-2 (13B)  &0.810 &0.782 
                    &0.824 &0.778 
                    &0.828 &0.780 
                    &0.836 &0.781 
                    &\textbf{0.838} &\textbf{0.783}  \\

& Ensemble &0.840 &0.786 
                &0.850 &0.787 
                &0.852 &0.790 
                &\textbf{0.853} &0.792 
                &\textbf{0.853} &\textbf{0.795}  \\

\midrule

%% Llama-2 (13B) results
\multirow{4}{*}{\textbf{Llama-2 (13B)}}
% \multirow{4}{*}{\rotatebox{90}{\textbf{Llama-2 (13B)}}}

& GPT3 &0.775 &0.752 
             &0.799 &0.754 
             &0.808 &\textbf{0.762} 
             &0.815 &0.761 
             &\textbf{0.819} &0.760  \\

& Llama-2 (7B) &0.757 &0.704 
                 &0.763 &\textbf{0.710} 
                 &0.764 &0.701 
                 &0.767 &0.702 
                 &\textbf{0.769} &0.699  \\

& Llama-2 (13B)  &0.770 &0.729 
                        &0.779 &0.731 
                        &0.786 &0.732 
                        &0.789 &\textbf{0.736 }
                        &\textbf{0.790} &0.734 \\

& Ensemble  &0.819 &0.754 
                 &0.821 &0.758 
                 &0.823 &0.758 
                 &0.823 &\textbf{0.759 }
                 &\textbf{0.824} &0.755 \\

\bottomrule
\end{tabular}
}
\caption{Results for the Movies dataset. Results are reported for different k values, ranging from 1 to 5, average score. The highest AUC and B-ACC values for each row are presented in bold.}
\label{Tab:k_results_movies_avg}
\end{table*}

\begin{figure*}[t]
\includegraphics[width=1.0\linewidth]{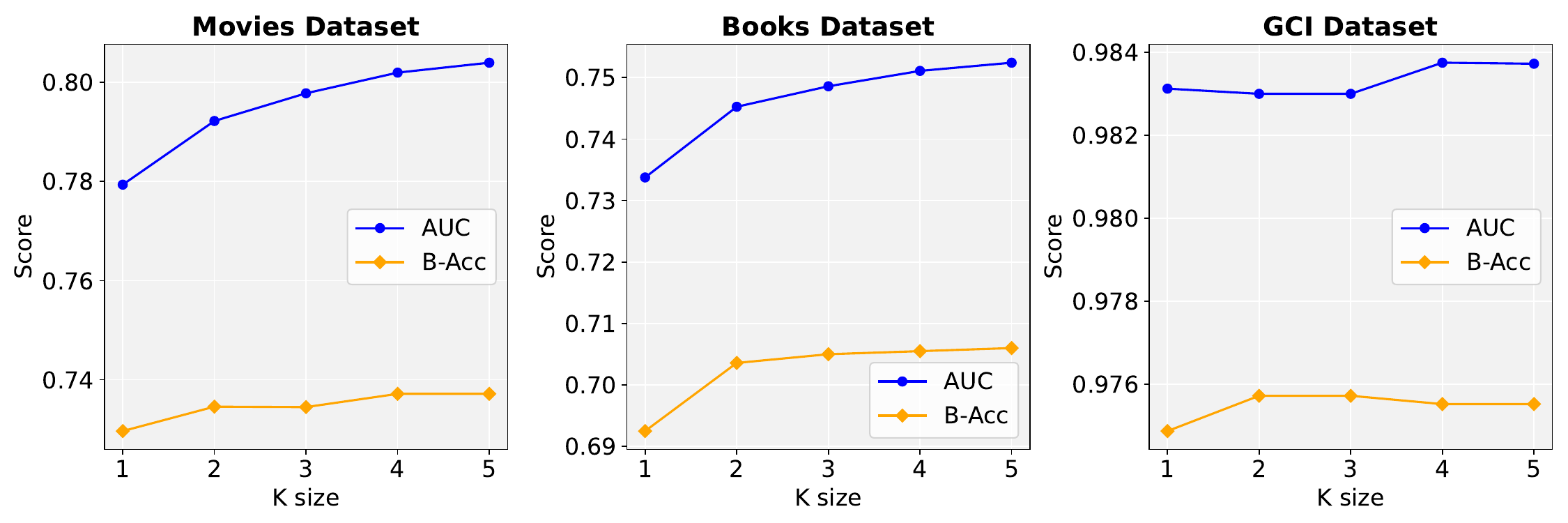}
\centering
\caption{
The average AUC and B-Acc scores across Movies, Books, and GCI datasets, per different K values (1-5). 
}
\label{fig:k_avg_curve}
\end{figure*}

% movies temp
\begin{table}[t!]

\centering
\resizebox{0.49\textwidth}{!}{

\begin{tabular}{@{}l@{~}l@{~}c@{~}c@{~}c@{~}c@{~}}
\toprule 
& & \multicolumn{2}{c}{Same temp}& \multicolumn{2}{c}{Variable temp}\\
\cmidrule(lr){3-4}
\cmidrule(lr){5-6}

\textbf{$\forwardModel$ }&  \textbf{$\backwardModel$ }&  
\textbf{\  AUC\ \ }&  \textbf{\ B-ACC\ \ }&
\textbf{\  AUC\ \ }&  \textbf{\ B-ACC\ \ }\\
\midrule

%% GPT 3 results
\multirow{4}{*}{GPT3}
&GPT3   &0.790 & 0.721 &\textbf{0.817} & \textbf{0.739}  \\
&Llama-2 (7B) &0.734 &0.631 &\textbf{0.751} & \textbf{0.639} \\
&Llama-2 (13B) &0.787 &\textbf{0.697} &\textbf{0.789} &0.695 \\
&Ensemble & 0.815 & 0.695 &\textbf{0.818} &\textbf{0.699} \\
\midrule

%% Llama-2 (7B) results
\multirow{4}{*}{Llama-2 (7B)}
&GPT3  &0.798 & 0.780 &\textbf{0.824} & \textbf{0.786}  \\
&Llama-2 (7B) &\textbf{0.830} &\textbf{0.766} &0.823 & 0.750 \\
&Llama-2 (13B) &\textbf{0.838} &\textbf{0.783} &0.828 &0.775 \\
&Ensemble  & 0.853 & 0.795 &\textbf{0.874} &\textbf{0.813} \\
\midrule

%% Llama-2 (13B) results
\multirow{4}{*}{Llama-2 (13B)}

&GPT3   &\textbf{0.819} & \textbf{0.760} &0.806 & 0.753  \\
&Llama-2 (7B) &0.769 &0.699 &\textbf{0.788} & \textbf{0.706} \\
&Llama-2 (13B) &0.790 &0.734 &\textbf{0.801} &\textbf{0.746} \\
&Ensemble  & 0.824 & 0.755 &\textbf{0.842} &\textbf{0.773} \\

\bottomrule
\bottomrule
\multicolumn{2}{c}{Average} & 0.803 & 0.734 &\textbf{0.813} &\textbf{0.739} \\
\bottomrule

\end{tabular}
}
\caption{Results for Movies dataset with same temperature and variable temperature.}
\label{Tab:results_movies_temperature}

\end{table}

\subsection{Ablation and hyper-parameter analysis}
\label{sec:ablation}
We conduct an ablation study to examine the impact of the multiple $K$ backward processes performed in \methodName{} (Alg.\ref{algo:qa_verification} line 4), the effectiveness of the variable temperature values (Eq.\eqref{eq:temperature}), and the importance of the average function in Eq.\ref{eq:avg}.

\paragraph{Variable $K$ values}
The performance of \methodName{} with various values of $K$ is evaluated on the Movies, Books, and GCI datasets, and the results are reported in Tab.~\ref{Tab:k_results_movies_avg}, and Tab. \ref{Tab:k_results_books_full}, \ref{Tab:k_results_gci_full} from the supplementary, respectively. Specifically, in this study, we evaluate \methodName{} with $K$ taking values in the range $[1,...,5]$ (higher $K$ values can be considered at the expense of more compute power). 
The tables reveal that utilizing $K>1$ in the backward step is crucial in all three experiments. Notably, the best results are consistently obtained with higher $K$ value, where $K=5$ takes the lead in the majority of cases. Therefore, we hypothesize that increasing the value of $K$ could potentially enhance the results, albeit at the expense of additional computational resources.
In addition, we observe that the ensemble of all three models (GPT-3, Llama-2 (7B), and Llama-2 (13B)) yielded the highest performance across all $K$ values. This suggests once again that combining recovery scores from multiple models enhances hallucination detection.

Fig.~\ref{fig:k_avg_curve} depicts the enhancements arising from the different values of $K$, shown for each dataset separately, and reported with both AUC and B-ACC metrics. 
Each data point represents the average result across all three forward LLMs along with all their corresponding backward LLMs (i.e. the average of each column in tables \ref{Tab:k_results_movies_avg},\ref{Tab:k_results_books_full} and \ref{Tab:k_results_gci_full}). As can be seen, the data reveals a consistent trend wherein the cumulative improvements exhibit a proportional relationship with the size of $K$.

\paragraph{Variable temperatures}
\label{sec:temp}
We extend our investigation to varying temperatures for the backward process. 
For each index $i \in \text{range}(K)$, the \methodName{} method utilizes a variable temperature $T_i$ as defined in Eq.\eqref{eq:temperature}. 
This temperature adjustment aimed to augment the creativity and stochastic aspects of the backward model $\backwardModel$ throughout the query reconstruction process, fostering the generation of a more diverse set of reconstructed queries. 
Tab.~\ref{Tab:results_movies_temperature}, and Tab.~\ref{Tab:results_books_temperature_full}, \ref{Tab:results_gci_temperature_full} from the supplementary Sec.\ref{sup:varTemp}, present the results of \methodName{} with $K=5$, when using the same temperature through all the backward processes versus using variable temperatures, as proposed in \methodName{}. As can be seen, the variable temperature improves the results across most experiments in the Movies datasets, while yielding on-par performance in the Books and GCI datasets (see Tab.\ref{Tab:results_books_temperature_full}, \ref{Tab:results_gci_temperature_full}). 
We hypothesize that the introduction of variable temperature, generating reconstructions with diverse levels of creativity, can be particularly helpful in mitigating instances of mode "collapse", in which certain backward models consistently generate identical reconstructions. In such cases, the incorporation of variable temperature becomes more important. The proposed method utilizes a diverse range of reconstructions. When the vast majority of these diverse reconstructions closely align with the original query, it signifies robust backward processes, better reflecting a non-hallucinated answer and consequently leading to improved accuracy scores. 
More ablative experiments related to the choices made in Eq.~\ref{eq:avg} can be found in the sup. Sec.\ref{sup:Average-Max}.

\section{Conclusion}

In this paper, we investigate the pressing issue of hallucinations in large language models.
We introduced \methodName{}, a novel method designed for detecting hallucinations in few-shot settings. 
Our work contributes to the ongoing dialogue on the responsible use of AI-powered language models, offering a method that contributes to the reliability of LLM in diverse real-world applications. 

As a future work, we would like to extend the method to Retrieval Augmented Generation settings, where a query is provided with a retrieved context, and the task is to generate an answer based on the provided information in the context. 
\section{Limitations}
\label{sec:limitations}

Throughout our study, we encountered several noteworthy limitations:
(1) Source and Target Domain with Many-to-One Mapping:
Generated answers associated with multiple different queries pose challenges in verification with \methodName{}. The backward process can reconstruct a diverse set of query candidates, deviating from the original query.
(2) Hallucinating Back and Forth:
Instances were observed where a single backward process by the same LLM model, which hallucinated an answer, could reconstruct the same query. This severe hallucination indicates a symmetric mapping between a query and a hallucinated answer, implying hallucinations in both directions. We observed a mitigation of this issue when employing an ensemble of models.
(3) Detecting Hallucinations in Semi-Truth Answers:
Identifying hallucinations in semi-truth answers proves more challenging. In some cases, the model only hallucinated a small portion of the answer (e.g., generating an entire cast of a movie with one additional actor not part of the movie). \methodName{} was able to recover the original movie, failing to detect the low-severity hallucination within the answer.


\bibliographystyle{acl_natbib}

\newpage

\appendix
% \label{sec:sample:appendix}
\begin{center}
{\LARGE Supplementary Appendices}
%\footnote{Put here for the reader's convenience.}}
\end{center}

\section{More results}
\label{sup:ablation}

\subsection{The Average-Max Analysis}
\label{sup:Average-Max}
For each query, we obtain a list of size K containing cosine similarity scores, representing the similarity between the generated query and the original query. To ascertain the closeness of the generated query to the original, we employ two strategies: the maximum (Max) and the average (Average) of the K cosine similarity scores. Notably, in the majority of cases, the average outperformed the maximum, indicating a more robust measure for query similarity. We attribute this observation to the sensitivity of the maximum strategy to noisy cosine similarity scores, which tend to be prevalent in certain instances. Our analysis suggests that the average is a more reliable indicator of query similarity, providing a stable and effective metric for the evaluation of our reconstruction approach.
Full results with avg/max are presented in tables \ref{Tab:results_movies_temperature_full}, \ref{Tab:results_books_temperature_full}, \ref{Tab:results_gci_temperature_full}, \ref{Tab:k_results_movies_full}, \ref{Tab:k_results_books_full} and \ref{Tab:k_results_gci_full}.

\subsection{More results for variable $K$ values}
\label{sup:variable_k}
Tab.~\ref{Tab:k_results_movies_full}, \ref{Tab:k_results_books_full} and \ref{Tab:k_results_gci_full}. depicts the performance of InterrogateLLM on the books and GCI datasets, with variable $K$ values. As can be seen, higher values of $K$ result in enhanced accuracy.

\subsection{More variable temperature results}
\label{sup:varTemp}
Tab.~\ref{Tab:results_movies_temperature_full}, \ref{Tab:results_books_temperature_full} and \ref{Tab:results_gci_temperature_full} depicts the performance of the \methodName{} method with and without variable temperature values.

\section{More information about the datasets}
\label{sec:samples}
\subsection{Movies Dataset Query-Answer Examples}
We provide query-answer examples from the Movies dataset, where each query seeks information about the cast of a specific movie:
\begin{itemize}
    \item Query-1:  "What actors played in the 1995 movie Jumanji?"  
    \item Answer-1: "The main cast included Robin Williams, Jonathan Hyde, Kirsten Dunst, Bradley Pierce, Bonnie Hunt, Bebe Neuwirth, David Alan Grier, Patricia Clarkson, Adam Hann-Byrd, Laura Bell Bundy, James Handy, Gillian Barber, Brandon Obray, Cyrus Thiedeke, Gary Joseph Thorup, Leonard Zola, Lloyd Berry, Malcolm Stewart, Annabel Kershaw, Darryl Henriques, Robyn Driscoll, Peter Bryant, Sarah Gilson, Florica Vlad, June Lion, Brenda Lockmuller."

    \item Query-2: "What actors played in the 2011 movie Kingdom Come?"
    \item Answer-2: "Selma Blair, Edward Burns, Bruce Campbell, Lizzy Caplan, Seymour Cassel, Don Cheadle, Joanne Cook, Rachael Leigh Cook, Tom Cook, Alan Cumming, Tom DiCillo, Drake Doremus."

    \item Query-3: "What actors played in the 2009 movie Inglourious Basterds?" 
    \item Answer-3: "The actors are Brad Pitt, Diane Kruger, Eli Roth, Mélanie Laurent, Christoph Waltz, Michael Fassbender, Daniel Brühl, Til Schweiger, Gedeon Burkhard, Jacky Ido, B.J. Novak, Omar Doom."
 
\end{itemize}

\subsection{Books Dataset Query-Answer Examples}
Query-Answer examples from the Books dataset, where queries inquire about the author and publication year of specific books:

\begin{itemize}
    \item Query-1: "Who is the author of the book Classical Mythology, and what year was it published?" 
    \item Answer-1: "The author is Mark P. O. Morford, and it was published in 2002."
    \item Query-2: "Who is the author of the book Decision in Normandy, and what year was it published?"
    \item Answer-2: "The author is Carlo D'Este, and it was published in 1991."
    \item Query-3: "Who is the author of the book Clara Callan, what year was it published?"
    \item Answer-3: "The author is Richard Bruce Wright, and it was published in 2001 by HarperFlamingo Canada."
\end{itemize}

\subsection{GCI Dataset Query-Answer Examples}
Query-Answer examples from the GCI dataset, where each query asks about the capital of a specific country:

\begin{itemize}
    \item Query-1: "What is the capital of France?" 
    \item Answer-1: "The capital is Paris."
    \item Query-2: "What is the capital of Japan?"
    \item Answer-2: "The capital is Tokyo."
    \item Query-3: "What is the capital of Australia?"
    \item Answer-3: "The capital is Canberra."
\end{itemize}

% movies temp
\begin{table}[ht!]

\centering
\resizebox{0.5\textwidth}{!}{

\begin{tabular}{@{}l@{~}l@{~}c@{~}c@{~}c@{~}c@{~}}
\toprule 
& & \multicolumn{2}{c}{Same temp}& \multicolumn{2}{c}{Variable temp}\\
\cmidrule(lr){3-4}
\cmidrule(lr){5-6}

\textbf{$\forwardModel$ }&  \textbf{$\backwardModel$ }&  
\textbf{\  AUC\ \ }&  \textbf{\ B-ACC\ \ }&
\textbf{\  AUC\ \ }&  \textbf{\ B-ACC\ \ }\\
\midrule

%% GPT 3 results
\multirow{8}{*}{GPT3}

&GPT3 (avg)  &0.790 & 0.721 &\textbf{0.817} & \textbf{0.739}  \\
&GPT3 (max)  &0.768 & 0.730 &\textbf{0.787} & \textbf{0.752}  \\

\cline{2-6}

&Llama-2 (7B) (avg)&0.734 &0.631 &\textbf{0.751} & \textbf{0.639} \\
&Llama-2 (7B) (max)&0.721 &0.669 &\textbf{0.726} & \textbf{0.690} \\

\cline{2-6}

&Llama-2 (13B) (avg)&0.787 &\textbf{0.697} &\textbf{0.789} &0.695 \\
&Llama-2 (13B) (max)&0.766 &0.725 &\textbf{0.772} &\textbf{0.732} \\

\cline{2-6}

&Ensemble  (avg)& 0.815 & 0.695 &\textbf{0.818} &\textbf{0.699} \\
&Ensemble  (max)& 0.786 & 0.741 &\textbf{0.798} &\textbf{0.756} \\

\midrule

%% Llama-2 (7B) results
\multirow{8}{*}{Llama-2 (7B)}
&GPT3 (avg)  &0.798 & 0.780 &\textbf{0.824} & \textbf{0.786}  \\
&GPT3 (max)  &0.758 & 0.765 &\textbf{0.776} & \textbf{0.768}  \\

\cline{2-6}

&Llama-2 (7B) (avg)&\textbf{0.830} &\textbf{0.766} &0.823 & 0.750 \\
&Llama-2 (7B) (max)&\textbf{0.814} &\textbf{0.781} &0.808 & 0.773 \\

\cline{2-6}

&Llama-2 (13B) (avg)&\textbf{0.838} &\textbf{0.783} &0.828 &0.775 \\
&Llama-2 (13B) (max)&\textbf{0.821} &\textbf{0.791} &0.802 &0.780 \\

\cline{2-6}

&Ensemble  (avg)& 0.853 & 0.795 &\textbf{0.874} &\textbf{0.813} \\
&Ensemble  (max)& 0.802 & 0.765 &\textbf{0.810} &\textbf{0.772} \\

\midrule

%% Llama-2 (13B) results
\multirow{8}{*}{Llama-2 (13B)}

&GPT3 (avg)  &\textbf{0.819} & \textbf{0.760} &0.806 & 0.753  \\
&GPT3 (max)  &\textbf{0.777} & \textbf{0.755} &0.769 & 0.748  \\

\cline{2-6}

&Llama-2 (7B) (avg)&0.769 &0.699 &\textbf{0.788} & \textbf{0.706} \\
&Llama-2 (7B) (max)&0.757 &0.728 &\textbf{0.772} & \textbf{0.738} \\

\cline{2-6}

&Llama-2 (13B) (avg)&0.790 &0.734 &\textbf{0.801} &\textbf{0.746} \\
&Llama-2 (13B) (max)&0.770 &0.739 &\textbf{0.777} &\textbf{0.748} \\

\cline{2-6}

&Ensemble  (avg)& 0.824 & 0.755 &\textbf{0.842} &\textbf{0.773} \\
&Ensemble  (max)& 0.763 & 0.733 &\textbf{0.792} &\textbf{0.756} \\
\bottomrule
\bottomrule
\multicolumn{2}{c}{Average  (avg)}& 0.803 & 0.734 &\textbf{0.813} &\textbf{0.739} \\
\multicolumn{2}{c}{Average  (max)}& 0.775 & 0.743 &\textbf{0.782} &\textbf{0.751} \\
\bottomrule

\end{tabular}
}
\caption{Results for Movies dataset, presenting results for constant and variable temperature, with both average and maximum scores.}
\label{Tab:results_movies_temperature_full}
\end{table}

% books temp
\begin{table}[ht!]

\centering
\resizebox{0.5\textwidth}{!}{

\begin{tabular}{@{}l@{~}l@{~}c@{~}c@{~}c@{~}c@{~}}

\toprule 
& & \multicolumn{2}{c}{Same temp}& \multicolumn{2}{c}{Variable temp}\\
\cmidrule(lr){3-4}
\cmidrule(lr){5-6}

\textbf{$\forwardModel$ }&  \textbf{$\backwardModel$ }&  
\textbf{\  AUC\ \ }&  \textbf{\ B-ACC\ \ }&
\textbf{\  AUC\ \ }&  \textbf{\ B-ACC\ \ }\\
\midrule

%% GPT 3 results
\multirow{8}{*}{GPT3}
&GPT3 (avg)  &0.698 & \textbf{0.675} &\textbf{0.709} & 0.673  \\
&GPT3 (max)  &0.685 & \textbf{0.670} &\textbf{0.694} & 0.667  \\

\cline{2-6}

&Llama-2 (7B) (avg)&0.640 &0.616 &\textbf{0.646} & 0.616 \\
&Llama-2 (7B) (max)&0.615 &0.619 &\textbf{0.632} & \textbf{0.625} \\

\cline{2-6}

&Llama-2 (13B) (avg)&0.675 &\textbf{0.642} &\textbf{0.684} &0.640 \\
&Llama-2 (13B) (max)&0.656 &0.643 &\textbf{0.669} &\textbf{0.648} \\

\cline{2-6}

&Ensemble  (avg)& 0.707 & 0.656 &\textbf{0.710} &0.656 \\
&Ensemble  (max)& 0.707 & 0.669 &\textbf{0.719} &\textbf{0.681} \\

\midrule

%% Llama-2 (7B) results
\multirow{8}{*}{Llama-2 (7B)}

&GPT3 (avg)  &0.821 & 0.777 &\textbf{0.828} & \textbf{0.787}  \\
&GPT3 (max)  &0.811 & 0.780 &\textbf{0.815} & \textbf{0.784}  \\

\cline{2-6}

&Llama-2 (7B) (avg)&0.761 &0.707 &0.761 & 0.707 \\
&Llama-2 (7B) (max)&0.744 &0.718 &\textbf{0.752} & \textbf{0.725} \\

\cline{2-6}

&Llama-2 (13B) (avg)&0.794 &0.730 &\textbf{0.795} &\textbf{0.734} \\
&Llama-2 (13B) (max)&0.783 &0.745 &\textbf{0.785} &\textbf{0.752} \\

\cline{2-6}

&Ensemble  (avg)& \textbf{0.824} & \textbf{0.769} &0.822 &0.761 \\
&Ensemble  (max)& \textbf{0.831} & \textbf{0.793} &0.827 &0.783 \\
\midrule

%% Llama-2 (13B) results
\multirow{8}{*}{Llama-2 (13B)}

&GPT3 (avg)  &0.799 & \textbf{0.757} &\textbf{0.804} & 0.754  \\
&GPT3 (max)  &0.792 & 0.758 &\textbf{0.797} & \textbf{0.763}  \\

\cline{2-6}

&Llama-2 (7B) (avg)&\textbf{0.743} &\textbf{0.686} &0.742 & 0.679 \\
&Llama-2 (7B) (max)&0.722 &0.696 &\textbf{0.731} & \textbf{0.707} \\

\cline{2-6}

&Llama-2 (13B) (avg)&0.771 &0.707 &0.771 &\textbf{0.709} \\
&Llama-2 (13B) (max)&0.754 &0.714 &\textbf{0.759} &\textbf{0.724} \\

\cline{2-6}

&Ensemble  (avg)& 0.802 & \textbf{0.739} &\textbf{0.807} &0.733 \\
&Ensemble  (max)& 0.808 & 0.765 &\textbf{0.817} &\textbf{0.774} \\
\bottomrule
\bottomrule
\multicolumn{2}{c}{Average  (avg)}& 0.752 & \textbf{0.705} &\textbf{0.756} &0.704 \\
\multicolumn{2}{c}{Average  (max)}& 0.742 & 0.714 &\textbf{0.747} &\textbf{0.719} \\
\bottomrule

\end{tabular}
}
\caption{Results for Books dataset, presenting results for constant and variable temperature, with both average and maximum scores.}
\label{Tab:results_books_temperature_full}
\end{table}

% gci temp
\begin{table}[ht!]

\centering
\resizebox{0.5\textwidth}{!}{

\begin{tabular}{@{}l@{~}l@{~}c@{~}c@{~}c@{~}c@{~}}

\toprule 
& & \multicolumn{2}{c}{Same temp}& \multicolumn{2}{c}{Variable temp}\\
\cmidrule(lr){3-4}
\cmidrule(lr){5-6}

\textbf{$\forwardModel$ }&  \textbf{$\backwardModel$ }&  
\textbf{\  AUC\ \ }&  \textbf{\ B-ACC\ \ }&
\textbf{\  AUC\ \ }&  \textbf{\ B-ACC\ \ }\\
\midrule

%% GPT 3 results
\multirow{8}{*}{GPT3}
&GPT3 (avg)  &- & 0.994 &- &0.994  \\
&GPT3 (max)  &- & 0.994 &- &0.994  \\

\cline{2-6}

&Llama-2 (7B) (avg)  &- & 0.983 &- &0.983  \\
&Llama-2 (7B) (max)  &- & 0.983 &- &0.983  \\

\cline{2-6}

&Llama-2 (13B) (avg) &- & 0.983 &- &0.983  \\
&Llama-2 (13B) (max) &- & 0.983 &- &0.983  \\

\cline{2-6}

&Ensemble (avg)  &- & 0.983 &- &0.983  \\
&Ensemble (max) &- & 0.983 &- &0.983  \\

\midrule

%% Llama-2 (7B) results
\multirow{8}{*}{Llama-2 (7B)}

&GPT3 (avg)  &\textbf{0.969} &\textbf{0.972} &0.965 &0.952  \\
&GPT3 (max)  &\textbf{0.968} & \textbf{0.972} &0.964 &0.952  \\

\cline{2-6}

&Llama-2 (7B) (avg)  &\textbf{0.974} &0.957 &0.959 &\textbf{0.958}  \\
&Llama-2 (7B) (max)  &\textbf{0.976} &0.961 &0.960 &\textbf{0.962}  \\

\cline{2-6}

&Llama-2 (13B) (avg)  &\textbf{0.977} &0.959 &0.969 &\textbf{0.960}  \\
&Llama-2 (13B) (max)  &\textbf{0.977} &0.959 &0.971 &\textbf{0.967}  \\

\cline{2-6}

&Ensemble (avg)  &\textbf{0.963} &\textbf{0.951} &0.951 &0.948 \\
&Ensemble (max) &0.944 &\textbf{0.944} &\textbf{0.949} &0.941 \\
\midrule

%% Llama-2 (13B) results
\multirow{8}{*}{Llama-2 (13B)}
&GPT3 (avg)  &0.986 &0.982 &\textbf{0.989} &0.982  \\
&GPT3 (max)  &0.971 & \textbf{0.978} &\textbf{0.983} &0.977  \\

\cline{2-6}

&Llama-2 (7B) (avg)  &1.000 &1.000 &1.000 &1.000  \\
&Llama-2 (7B) (max)  &1.000 &1.000 &1.000 &1.000  \\

\cline{2-6}

&Llama-2 (13B) (avg)  &\textbf{1.000} &0.991 &0.995 &0.991 \\
&Llama-2 (13B) (max)  &\textbf{0.998} &0.978 &0.983 &\textbf{0.986}  \\

\cline{2-6}

&Ensemble (avg)  &\textbf{1.000} &\textbf{0.991} &0.992 &0.964 \\
&Ensemble (max) &\textbf{0.987} &\textbf{0.986} &0.983 &0.967 \\

\bottomrule
\bottomrule
\multicolumn{2}{c}{Average  (avg)}& \textbf{0.983} & 0.974 &0.977 &0.974 \\
\multicolumn{2}{c}{Average  (max)}& \textbf{0.977} & \textbf{0.976} &0.974 &0.974 \\
\bottomrule
\end{tabular}
}
\caption{Results for GCI dataset, presenting results for constant and variable temperature, with both average and maximum scores.}
\label{Tab:results_gci_temperature_full}
\end{table}

% movies k=1 - k=5 results
\begin{table*}[ht!]

\centering
\resizebox{1.0\textwidth}{!}{

\begin{tabular}{@{}l@{~}l@{~}c@{~}c@{~}c@{~}c@{~}c@{~}c@{~}c@{~}c@{~}c@{~}c@{~}c@{~}c@{}}
\toprule 
& & \multicolumn{2}{c}{k=1}& \multicolumn{2}{c}{k=2}& \multicolumn{2}{c}{k=3} & \multicolumn{2}{c}{k=4}& \multicolumn{2}{c}{k=5}\\
\cmidrule(lr){3-4}
\cmidrule(lr){5-6}
\cmidrule(lr){7-8}
\cmidrule(lr){9-10}
\cmidrule(lr){11-12}
\textbf{$\forwardModel$ }&  \textbf{$\backwardModel$ }& 
\textbf{\  AUC\ \ }&  \textbf{\ B-ACC\ \ }&
\textbf{\  AUC\ \ }&  \textbf{\ B-ACC\ \ }&
\textbf{\  AUC\ \ }&  \textbf{\ B-ACC\ \ }&
\textbf{\  AUC\ \ }&  \textbf{\ B-ACC\ \ }&
\textbf{\  AUC\ \ }&  \textbf{\ B-ACC\ \ }\\
\midrule

%% GPT 3 results
\multirow{8}{*}{GPT3}
& GPT3 (max) &0.755 &0.710 
            &0.765 &0.724 
            &0.768 & 0.730 
            &0.767 &0.729 
            &0.768 & 0.730  \\
            
& GPT3 (avg)&0.755 &0.710 
            &0.773 &0.722 
            &0.782 & 0.719 
            &0.786 &0.720 
            &0.790 & 0.721  \\
\cline{2-12}
& Llama-2 (7B) (max)&0.701 &0.633 
                    &0.714 &0.650 
                    &0.714 &0.659 
                    &0.718 &0.664 
                    &0.721 &0.669 \\
                    
& Llama-2 (7B) (avg) &0.701 &0.633 
                    &0.721 &0.641 
                    &0.727 &0.635 
                    &0.732 &0.638 
                    &0.734 &0.631 \\
\cline{2-12}
& Llama-2 (13B) (max)&0.756 &0.688 
                    &0.761 &0.707 
                    &0.764 &0.715  
                    &0.765 &0.721 
                    &0.766 &0.725 \\
                    
& Llama-2 (13B) (avg)&0.756 &0.688 
                    &0.772 &0.696 
                    &0.779 &0.698 
                    &0.783 &0.696
                    &0.787 &0.697 \\

\cline{2-12}
 & Ensemble (max) &0.782 &0.736 
                &0.778 &0.742 
                &0.785 &0.744  
                &0.787 &0.745 
                &0.786 &0.741 \\
                
 & Ensemble (avg) & 0.796 & 0.690 
                &0.803 &0.694 
                &0.811 &0.694 
                &0.814 &0.695 
                &0.815 &0.695 \\
\midrule

%% Llama-2 (7B) results
\multirow{8}{*}{Llama-2 (7B)}
& GPT3 (max) &0.775 &0.774 
            &0.767 &0.775 
            &0.761 & 0.769 
            &0.758 & 0.766
            &0.758 & 0.765 \\
            
& GPT3 (avg) &0.775 &0.774 
            &0.786 &0.778 
            &0.788 & 0.776
            &0.794 & 0.782 
            &0.798 & 0.780  \\
            
\cline{2-12}
& Llama-2 (7B)  (max)&0.798 &0.754 
                    &0.808 &0.775 
                    &0.812 &0.778 
                    &0.818 &0.782 
                    &0.814 &0.781  \\
                    
& Llama-2 (7B)  (avg)&0.798 &0.754 
                    &0.815 &0.766 
                    &0.825 &0.757 
                    &0.831 &0.760 
                    &0.830 &0.766  \\
\cline{2-12}
& Llama-2 (13B) (max) &0.810 &0.782 
                    &0.812 &0.781 
                    &0.814 &0.785 
                    &0.822 &0.791 
                    &0.821 &0.791  \\
                    
& Llama-2 (13B) (avg) &0.810 &0.782 
                    &0.824 &0.778 
                    &0.828 &0.780 
                    &0.836 &0.781 
                    &0.838 &0.783  \\
\cline{2-12}
& Ensemble (max)  &0.810 &0.782 
                    &0.810 &0.776 
                    &0.810 &0.773 
                    &0.809 &0.770 
                    &0.802 &0.765  \\
                    
& Ensemble (avg) &0.840 &0.786 
                &0.850 &0.787 
                &0.852 &0.790 
                &0.853 &0.792 
                &0.853 &0.795  \\

\midrule

%% Llama-2 (13B) results
\multirow{8}{*}{Llama-2 (13B)}
 & GPT3 (max) &0.775 &0.752 
             &0.779 &0.755 
             &0.775 &0.753 
             &0.777 &0.752 
             &0.777 &0.755  \\
             
 & GPT3 (avg) &0.775 &0.752 
             &0.799 &0.754 
             &0.808 &0.762 
             &0.815 &0.716 
             &0.819 &0.760  \\
 
\cline{2-12}
& Llama-2 (7B) (max) &0.757 &0.704 
                    &0.758 &0.717 
                    &0.754 &0.721 
                    &0.753 &0.727 
                    &0.757 &0.728  \\

 & Llama-2 (7B) (avg) &0.757 &0.704 
                 &0.763 &0.710 
                 &0.764 &0.701 
                 &0.767 &0.702 
                 &0.769 &0.699  \\
\cline{2-12}
& Llama-2 (13B) (max) &0.770 &0.729 
                        &0.770 &0.732 
                        &0.770 &0.735 
                        &0.770 &0.736
                        &0.770 &0.739 \\
                        
& Llama-2 (13B) (avg) &0.770 &0.729 
                        &0.779 &0.731 
                        &0.786 &0.732 
                        &0.789 &0.736 
                        &0.790 &0.734 \\
\cline{2-12}
& Ensemble (max) &0.793 &0.765 
                &0.777 &0.751 
                &0.774 &0.743 
                &0.766 &0.739 
                &0.763 &0.733 \\
                
 & Ensemble (avg) &0.819 &0.754 
                 &0.821 &0.758 
                 &0.823 &0.758 
                 &0.823 &0.759 
                 &0.824 &0.755 \\

\bottomrule
\end{tabular}
}
\caption{Evaluation results for the Movies dataset across different k values (1 to 5), with average and maximum scores presented.}
\label{Tab:k_results_movies_full}
\end{table*}

% books k=1 - k=5 results
\begin{table*}[ht!]

\centering
\resizebox{1.0\textwidth}{!}{

\begin{tabular}{@{}l@{~}l@{~}c@{~}c@{~}c@{~}c@{~}c@{~}c@{~}c@{~}c@{~}c@{~}c@{~}c@{~}c@{}}
\toprule 
& & \multicolumn{2}{c}{k=1}& \multicolumn{2}{c}{k=2}& \multicolumn{2}{c}{k=3} & \multicolumn{2}{c}{k=4}& \multicolumn{2}{c}{k=5}\\
\cmidrule(lr){3-4}
\cmidrule(lr){5-6}
\cmidrule(lr){7-8}
\cmidrule(lr){9-10}
\cmidrule(lr){11-12}
\textbf{$\forwardModel$ }&  \textbf{$\backwardModel$ }& 
\textbf{\  AUC\ \ }&  \textbf{\ B-ACC\ \ }&
\textbf{\  AUC\ \ }&  \textbf{\ B-ACC\ \ }&
\textbf{\  AUC\ \ }&  \textbf{\ B-ACC\ \ }&
\textbf{\  AUC\ \ }&  \textbf{\ B-ACC\ \ }&
\textbf{\  AUC\ \ }&  \textbf{\ B-ACC\ \ }\\
\midrule

%% GPT 3 results
\multirow{8}{*}{GPT3}
& GPT3 (max)        &0.680 &0.657 
                    &0.683 &0.671 
                    &0.681 &0.669 
                    &0.682 &0.669 
                    &0.685 &0.670  \\
                    
& GPT3 (avg)        &0.680 &0.657 
                    &0.691 &0.673 
                    &0.692 &0.676 
                    &0.695 &0.681 
                    &0.698 &0.675  \\
\cline{2-12}
& Llama-2 (7B) (max)&0.626 &0.606 
                    &0.621 &0.615 
                    &0.613 &0.614 
                    &0.610 &0.616 
                    &0.609 &0.617  \\
                    
& Llama-2 (7B) (avg)&0.626 &0.606 
                    &0.635 &0.615 
                    &0.633 &0.614 
                    &0.634 &0.614 
                    &0.634 &0.615  \\
\cline{2-12}
& Llama-2 (13B) (max)&0.654 &0.623 
                    &0.654 &0.634
                    &0.659 &0.640
                    &0.660 &0.643
                    &0.656 &0.643 \\
                    
& Llama-2 (13B) (avg)&0.654 &0.623 
                    &0.665 &0.633
                    &0.670 &0.637
                    &0.673 &0.638
                    &0.675 &0.642 \\
\cline{2-12}
& Ensemble  (max)   &0.693 &0.668 
                    &0.696 &0.670
                    &0.698 &0.668
                    &0.703 &0.668
                    &0.707 &0.669 \\
                    
& Ensemble  (avg)   &0.696 &0.658 
                    &0.703 &0.663
                    &0.704 &0.657
                    &0.706 &0.655
                    &0.707 &0.656 \\
\midrule

%% Llama-2 (7B) results
\multirow{8}{*}{Llama-2 (7B)}

& GPT3 (max) &0.795 &0.757 
            &0.804 &0.771 
            &0.804 & 0.773 
            &0.809 & 0.777 
            &0.811 & 0.780 \\
            
& GPT3 (avg)&0.795 &0.757 
            &0.811 &0.772 
            &0.815 & 0.773 
            &0.820 & 0.774 
            &0.821 & 0.777  \\
            
\cline{2-12}
& Llama-2 (7B)  (max)&0.737 &0.686 
                    &0.744 &0.704 
                    &0.746 &0.712 
                    &0.743 &0.714 
                    &0.744 &0.718  \\
                    
& Llama-2 (7B)  (avg)&0.737 &0.686 
                    &0.754 &0.703 
                    &0.760 &0.708 
                    &0.760 &0.709 
                    &0.761 &0.707  \\
\cline{2-12}
& Llama-2 (13B) (max)&0.773 &0.720  
                    &0.778 &0.734  
                    &0.779 &0.738  
                    &0.781 &0.741  
                    &0.783 &0.745  \\
                    
& Llama-2 (13B) (avg)&0.773 &0.720  
                    &0.785 &0.729  
                    &0.791 &0.732  
                    &0.793 &0.731  
                    &0.794 &0.730  \\
                    
\cline{2-12}
& Ensemble   (max)&0.806 &0.777 
                &0.818 &0.787 
                &0.822 & 0.789 
                &0.827 &0.793 
                &0.831 & 0.793  \\
                
& Ensemble   (avg)&0.811 &0.766 
                &0.817 &0.768 
                &0.819 & 0.764 
                &0.822 &0.764 
                &0.824 & 0.769  \\
\midrule

%% Llama-2 (13B) results
\multirow{8}{*}{Llama-2 (13B)}
& GPT3   (max) &0.776 &0.733 
            &0.782 &0.745 
            &0.783 & 0.748
            &0.788 &0.755 
            &0.792 & 0.758  \\
            
& GPT3   (avg) &0.776 &0.733 
                &0.789 &0.750 
                &0.794 & 0.755 
                &0.797 &0.754 
                &0.799 & 0.757 \\
                
\cline{2-12}
& Llama-2 (7B) (max) &0.716 &0.674 
                        &0.721 &0.688 
                        &0.724 &0.695 
                        &0.722 &0.695 
                        &0.722 &0.696  \\
                        
& Llama-2 (7B) (avg) &0.716 &0.674 
                    &0.732 &0.689 
                    &0.740 &0.690 
                    &0.743 &0.690 
                    &0.743 &0.686  \\
\cline{2-12}
& Llama-2 (13B)  (max) &0.748 &0.694 
                        &0.749 &0.706 
                        &0.749 &0.709 
                        &0.751 &0.714 
                        &0.754 &0.714   \\
                        
& Llama-2 (13B)  (avg) &0.748 &0.694 
                        &0.761 &0.707 
                        &0.764 &0.703 
                        &0.769 &0.707 
                        &0.771 &0.707   \\
\cline{2-12}
& Ensemble (max) &0.788 &0.752 
                    &0.800 &0.765 
                    &0.803 &0.765 
                    &0.807 &0.766 
                    &0.808 &0.765 \\
                    
& Ensemble (avg) &0.793 &0.736 
                &0.800 &0.741 
                &0.801 &0.739 
                &0.801 &0.737 
                &0.802 &0.739 \\

\bottomrule
\end{tabular}
}
\caption{Evaluation results for the Books dataset across different k values (1 to 5), with average and maximum scores presented.}
\label{Tab:k_results_books_full}
\end{table*}

% gci k=1 - k=5 results
\begin{table*}[ht!]

\centering
\resizebox{1.0\textwidth}{!}{

\begin{tabular}{@{}l@{~}l@{~}c@{~}c@{~}c@{~}c@{~}c@{~}c@{~}c@{~}c@{~}c@{~}c@{~}c@{~}c@{}}
\toprule 
& & \multicolumn{2}{c}{k=1}& \multicolumn{2}{c}{k=2}& \multicolumn{2}{c}{k=3} & \multicolumn{2}{c}{k=4}& \multicolumn{2}{c}{k=5}\\
\cmidrule(lr){3-4}
\cmidrule(lr){5-6}
\cmidrule(lr){7-8}
\cmidrule(lr){9-10}
\cmidrule(lr){11-12}
\textbf{$\forwardModel$ }&  \textbf{$\backwardModel$ }& 
\textbf{\  AUC\ \ }&  \textbf{\ B-ACC\ \ }&
\textbf{\  AUC\ \ }&  \textbf{\ B-ACC\ \ }&
\textbf{\  AUC\ \ }&  \textbf{\ B-ACC\ \ }&
\textbf{\  AUC\ \ }&  \textbf{\ B-ACC\ \ }&
\textbf{\  AUC\ \ }&  \textbf{\ B-ACC\ \ }\\
\midrule

%% GPT 3 results
\multirow{8}{*}{GPT3}
& GPT3 (max) &- &0.994 
            &- &0.994 
            &- &0.994 
            &- &0.994 
            &- &0.994  \\
            
& GPT3 (avg) &- &0.994
            &- &0.994 
            &- & 0.994 
            &- &0.994 
            &- & 0.994  \\
            
\cline{2-12}
& Llama-2 (7B) (max) &- &0.983 
                    &- &0.983 
                    &- & 0.983 
                    &- &0.983 
                    &- & 0.983  \\

& Llama-2 (7B) (avg) &- &0.983 
                    &- &0.983 
                    &- &0.983 
                    &- &0.983 
                    &- & 0.983  \\
\cline{2-12}
& Llama-2 (13B) (max) &- &0.983 
                    &- &0.983 
                    &- & 0.983 
                    &- &0.983 
                    &- & 0.983  \\

& Llama-2 (13B) (avg) &- &0.983 
                    &- &0.983 
                    &- &0.983 
                    &- &0.983 
                    &- & 0.983  \\
\cline{2-12}
& Ensemble (max) &- &0.983 
                    &- &0.983 
                    &- & 0.983 
                    &- &0.983 
                    &- & 0.983  \\

& Ensemble (avg) &- &0.983 
                    &- &0.983 
                    &- &0.983 
                    &- &0.983 
                    &- & 0.983  \\
\midrule

%% Llama-2 (7B) results
\multirow{8}{*}{Llama-2 (7B)}
& GPT3  (max) &0.969 &0.968 
            &0.969 &0.968 
            &0.969 &0.972 
            &0.968 &0.972 
            &0.968 &0.972 \\
            
& GPT3  (avg) &0.969 &0.968 
            &0.969 &0.968 
            &0.969 &0.972 
            &0.970 &0.972 
            &0.969 &0.972  \\
            
\cline{2-12}

& Llama-2 (7B)  (max)&0.975 &0.957 
                    &0.976 &0.961 
                    &0.976 &0.961 
                    &0.976 &0.961 
                    &0.976 &0.961  \\
                    
& Llama-2 (7B)  (avg)&0.975 &0.957 
                    &0.975 &0.961 
                    &0.974 &0.961 
                    &0.974 &0.957 
                    &0.974 &0.957  \\
                    
\cline{2-12}
& Llama-2 (13B) (max)&0.977 &0.959 
                    &0.977 &0.959 
                    &0.977 &0.959 
                    &0.977 &0.959 
                    &0.977 &0.959  \\
                    
& Llama-2 (13B) (avg)&0.977 &0.959 
                    &0.977 &0.959 
                    &0.977 &0.959 
                    &0.977 &0.959 
                    &0.977 &0.959  \\
\cline{2-12}
& Ensemble   (max)&0.945 &0.944 
                    &0.944 &0.944 
                    &0.944 &0.944 
                    &0.944 &0.944 
                    &0.944 &0.944  \\
                    
& Ensemble   (avg)&0.964 &0.951 
                    &0.963 &0.951 
                    &0.964 &0.951 
                    &0.963 &0.951
                    &0.963 &0.951  \\
\midrule

%% Llama-2 (13B) results
\multirow{8}{*}{Llama-2 (13B)}
& GPT3   (max) &0.980 &0.982 
                &0.980 &0.982 
                &0.980 &0.982 
                &0.980 &0.982  
                &0.971 &0.978 \\
                
& GPT3   (avg) &0.980 &0.982 
                &0.980 &0.982 
                &0.980 &0.982 
                &0.986 &0.982 
                &0.986 &0.982 \\
\cline{2-12}
& Llama-2 (7B)  (max) &1.000 &1.000 
                    &1.000 &1.000  
                    &1.000 &1.000 
                    &1.000 &1.000  
                    &1.000 &1.000  \\
                    
& Llama-2 (7B)  (avg)&1.000 &1.000 
                    &1.000 &1.000  
                    &1.000 &1.000 
                    &1.000 &1.000  
                    &1.000 &1.000  \\
\cline{2-12}
& Llama-2 (13B)  (max) &1.000&0.991 
                        &1.000&0.991 
                        &0.998 & 0.982 
                        &0.998 & 0.982 
                        &0.998 & 0.978\\
                        
& Llama-2 (13B)  (avg) &1.000&0.991 
                    &1.000&0.991 
                    &1.000&0.991 
                    &1.000&0.991 
                    &1.000&0.991\\
\cline{2-12}
& Ensemble (max) &0.987 &0.991 
                &0.987 &0.991 
                &0.987 &0.991 
                &0.987 &0.991 
                &0.987 &0.986 \\

& Ensemble (avg) &1.000&0.991 
                &1.000 &0.995 
                &1.000 &0.991 
                &1.000 &0.991 
                &1.000 &0.991\\

\bottomrule
\end{tabular}
}
\caption{Evaluation results for the GCI dataset across different k values (1 to 5), with average and maximum scores presented.}
\label{Tab:k_results_gci_full}
\end{table*}

\end{document}